\documentclass[sigconf]{acmart}

\AtBeginDocument{%
  \providecommand\BibTeX{{%
    \normalfont B\kern-0.5em{\scshape i\kern-0.25em b}\kern-0.8em\TeX}}}

\acmSubmissionID{979}

\usepackage{verbatim}
\usepackage{graphicx}
\usepackage[ruled,linesnumbered,lined,boxed,commentsnumbered]{algorithm2e}
\let\oldnl\nl% Store \nl in \oldnl
\SetEndCharOfAlgoLine{}
\newcommand{\nonl}{\renewcommand{\nl}{\let\nl\oldnl}}% Remove line number for one line

\usepackage{xcolor}
\usepackage{caption}
\usepackage{subcaption}

\SetAlFnt{\footnotesize}

\title{Evolution of a Web-Scale Near Duplicate \\ Image Detection System}
\author{Andrey Gusev* and Jiajing Xu*}
\affiliation{%
  \institution{Pinterest}
  \city{San Francisco}
  \state{CA}
}
\email{{andreygusev,jiajing}@pinterest.com}

\renewcommand{\shortauthors}{Gusev and Xu}
\renewcommand{\midrule}{\specialrule{.4pt}{2pt}{0pt}}
\renewcommand{\bottomrule}{\specialrule{.8pt}{0pt}{2pt}}

\setcopyright{iw3c2w3}
\begin{document}
%\ninept
%

% -----------------------------------------------------------
\begin{abstract}

Detecting near duplicate images is fundamental to the content ecosystem of photo sharing web applications. However, such a task is challenging when involving a web-scale image corpus containing billions of images. In this paper, we present an efficient system for detecting near duplicate images across 8 billion images. Our system consists of three stages: candidate generation, candidate selection, and clustering. We also demonstrate that this system can be used to greatly improve the quality of recommendations and search results across a number of real-world applications.

In addition, we include the evolution of the system over the course of six years, bringing out experiences and lessons on how new systems are designed to accommodate organic content growth as well as the latest technology. Finally, we are releasing a human-labeled dataset of \textasciitilde 53,000 pairs of images introduced in this paper.

\end{abstract}

\keywords{near-duplicate detection, recommendation systems, locality sensitive hashing, transfer learning, clustering}

\copyrightyear{2020}
\acmYear{2020}
\acmConference[WWW '20]{Proceedings of The Web Conference 2020}{April 20--24, 2020}{Taipei, Taiwan}
\acmBooktitle{Proceedings of The Web Conference 2020 (WWW '20), April 20--24, 2020, Taipei, Taiwan}
\acmPrice{}
\acmDOI{10.1145/3366423.3380031}
\acmISBN{978-1-4503-7023-3/20/04}

\maketitle
{\let\thefootnote\relax\footnotetext{*Both authors contributed equally.}}

% -----------------------------------------------------------
\section{Introduction}
\label{sec:intro}

% \colorbox{yellow!60}{Explain, pin, image, meta data}

\begin{figure}[H]
% \begin{minipage}[b]{1.0\linewidth}
  \centering
%  \centerline{\includegraphics[width=7.5cm]{figures/examples_of_neardups.png}}
  \centerline{\includegraphics[width=8.5cm]{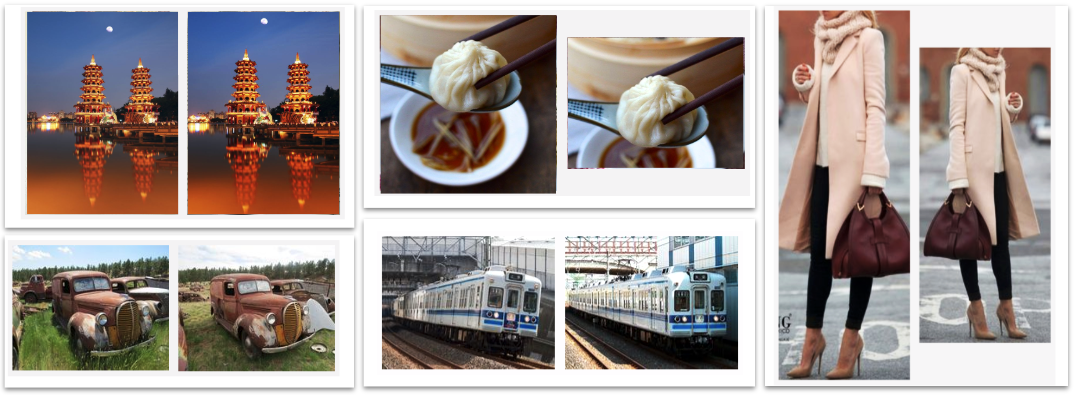}}
\vspace{-.3cm}
% \end{minipage}
\caption{Examples of near duplicate image pairs.}
\label{fig:neardup-def}
  \vspace{-.6cm}
\end{figure}

Near duplicate (near-dupe) image detection is a task of clustering exact and near identical images. There is no universal precise definition of near-dupe as it is very application platform specific. Typically, we refer to the same image with pixel-level modifications (e.g. cropping, flipping, scaling, rotating), or with a different perspective as a near-dupe image.  Figure \ref{fig:neardup-def} shows some examples of near-dupe image pairs.

\begin{figure}[H]
  \begin{subfigure}[b]{0.45\columnwidth}
    \includegraphics[width=\linewidth]{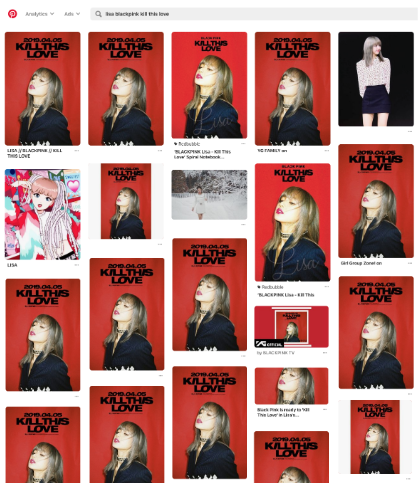}
    \caption{Without de-dupe filter}
    \label{fig:search-dedupe-before}
  \end{subfigure}
  \hfill %%
  \begin{subfigure}[b]{0.45\columnwidth}
    \includegraphics[width=\linewidth]{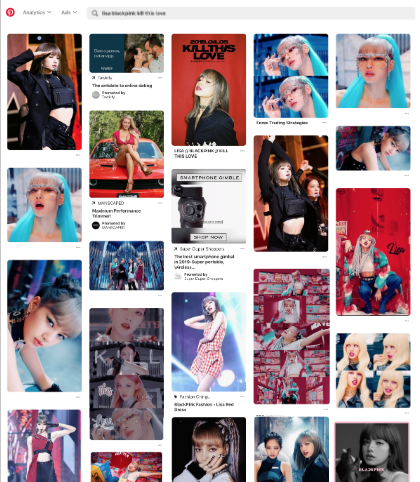}
    \caption{With de-dupe filter}
    \label{fig:search-dedupe-after}
  \end{subfigure} 
  \vspace{-.3cm}
  \caption{Results for search query "lisa blackpink kill this love" on Pinterest: (a) without image de-dupe filter (undesired user experience) (b) with image de-dupe filter.}
  \label{fig:search-dedupe}
\end{figure}

Near-dupe image detection is fundamental to photo-sharing web applications, such as Pinterest. For example, users expect to see relevant and diverse image search results. An excessive amount of near-dupe images would lead to an undesired user experience, as shown in Figure \ref{fig:search-dedupe-before}. As another example, recommendation systems can benefit from less noisy label and training data by aggregating meta data from near-dupe images.

%allows downstream systems to greatly improve relevance and density of content. 
% Excessive amount of near-dupe images would lead to an undesired user experience on search result pages, illustrated in Figure \ref{fig:search-dedupe-before}.  On the other hand, recommendation systems can benefit from aggregating meta data from near-dupe images.  Therefore, efficient and effective near-dupe image detection is fundamental to the health of Pinterest content ecosystem.

However, near-dupe image detection remains a challenging problem for a web-scale image corpus, because it is difficult to (1) capture the criteria of near-dupe and build a high-precision classifier, (2) scale to billions of images, and (3) process new images incrementally. Prior web-scale efforts \cite{liu2007clustering,wang2013duplicate,gong2015web,kim2015near}  have mainly focused on one of the above problems. But there has been limited work that addresses all of the above problems.

\begin{figure*}[htb]
    \begin{subfigure}[b]{1.5\columnwidth}
      \includegraphics[width=\linewidth]{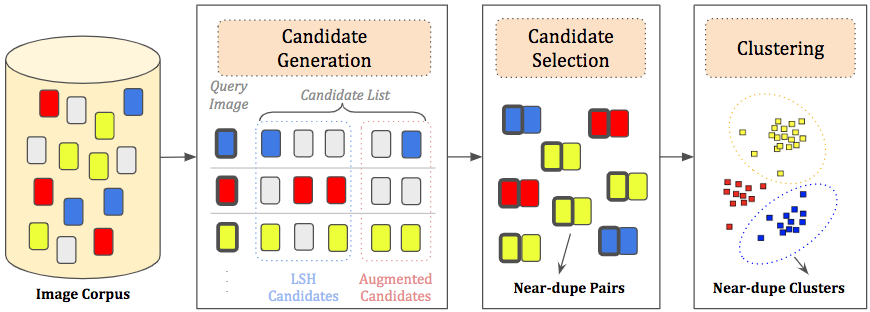}
      \caption{}
      \label{fig:system-diagram}
    \end{subfigure}
    \hspace{1em}
    \begin{subfigure}[b]{0.5\columnwidth}
      \includegraphics[width=\linewidth]{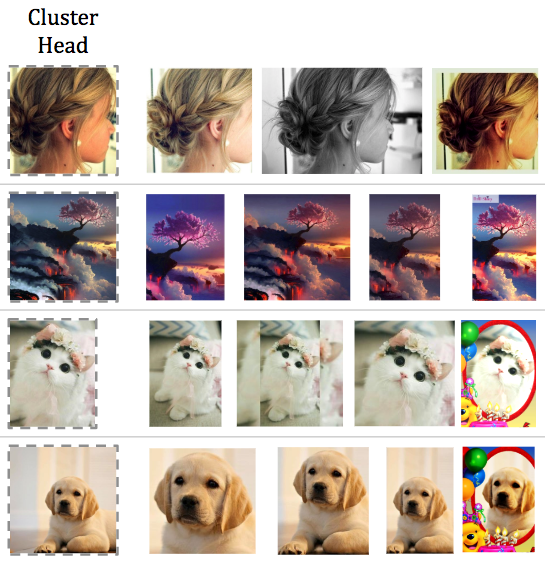}
      \caption{}
      \label{fig:neardup-cluster-example}
    \end{subfigure}
  \vspace{-.5cm}
  \caption{(a) System Diagram (b) Output of the near-dupe image detection system are near-dupe clusters.  Each row represents one near-dupe cluster, with the cluster head in dotted line and a few examples from the cluster.}
  \vspace{-.3cm}
\end{figure*}

This paper presents the design of our near-dupe image detection system, and shares our approach to handling the above three challenges. 
For the first challenge, we create a gold standard dataset using a crowdsourcing platform and train an initial classifier to optimize for precision using this dataset;
we then bootstrap over synthetic labels produced by the initial classifier and train a more powerful classifier based on visual embeddings.
For the second challenge, we map the visual embeddings into locality sensitive hashing (LSH) \cite{indyk1998approximate} terms and implement batch LSH search efficiently on Spark.
% achieving a 13x speedup over our previous generation system
For the third challenge, we develop an incremental update algorithm to process a new batch of images on a daily basis.

Our system detects near-dupe images across 8 billion images, the largest dataset to the best of our knowledge. Our method has three stages: candidate generation using a batch LSH algorithm, candidate selection through transfer learning over visual embeddings, and near-dupe clustering with incremental updates. 
We describe how to apply near-dupe signals to make a big impact on real-world applications such as Pinterest. We also include how to aggregate meta data in a general recommendation system using near-dupe signals. We share the lessons learned from the evolution of the system over the past six years, including methods that could improve efficiency.

The rest of the paper is organized as follows: Sections \ref{sec:related-work} and \ref{sec:system-overview} survey related works and give a brief overview of the near-dupe image detection systems built over the years. Section \ref{sec:methods} presents in detail the candidate generation, candidate selection, and clustering components of the latest system. Sections \ref{sec:experiments} and \ref{sec:applications} describe how near-dupe signals can be used to enhance Pinterest experiences. Section \ref{sec:discussions} shares what we have learned over the system's evolution.

% -----------------------------------------------------------
\section{Related Work}
\label{sec:related-work}

\subsection{Image Representation and Candidate Generation}  Previous work represents images by global descriptor \cite{liu2007clustering,lin2008edge, yue2011content}, local descriptor \cite{liu2010svd, li2013efficient, kim2015near}, bag-of-visual-words (BOW) \cite{chum2009geometric,chum2009large}, and a hybrid of global and local descriptor \cite{wang2013duplicate,zhou2017fast}.  Global descriptor is easy to compute but is vulnerable to heavy editing such as cropping.  Local descriptor and BOW-based methods are great alternatives to address this issue \cite{chum2009geometric,chum2009large, zhou2017_7546839, yao2014near,liu2015variable}.  The BOW model, based on local descriptor, is favorable for efficiency and scalability.  One challenge for the BOW model is that global context is lost in the representation. For the task of detecting near-dupe copies, an overlapping region-based global context descriptor is developed to increase the robustness of the features \cite{zhou2017_7546839}.  In recent years, deep convolutional neural networks (CNN) have been extremely successful for computer vision applications and have become the dominant approach for classification \cite{krizhevsky2012imagenet,simonyan2014very} and object detection \cite{girshick2014rich, erhan2014scalable}.  CNN have also been used for the task of near-dupe detection \cite{su2018joint, hu2018_8362942, zhang2018learning}.  However, training an end-to-end CNN can be time-consuming and costly, and it also comes with extra computation and storage cost for the task-specific image features.  For web-scale application, it is more desirable to reuse visual embeddings that are already available.  In this work, each image is represented by visual embeddings trained for other computer vision tasks in \cite{jing2015visual}.

For candidate generation, web-scale efforts use variants of approximate nearest neighbor search (aKNN) \cite{liu2007clustering} and min-hash \cite{chum2008near} to generate cluster seeds and grow the clusters \cite{wang2013duplicate,kim2015near}.  LSH, an aKNN method that maps high-dimensional image representations into a low-dimensional space to produce binary hash code, has recently become a popular indexing method for large-scale datasets \cite{chum2008near,chum2009large,lee2010partition,wang2017general,hu2018_8362942}.  This paper adopts the LSH method and optimizes the implementation extensively on Spark. We discuss in Section \ref{sec:discussions} that task-agnostic LSH works well as a candidate generator.

\subsection{Near-dupe Classifier and Web-scale System}
Prior web-scale near-dupe detection systems use a decision stump as the near-dupe classifier:  \cite{liu2007clustering} is based on global descriptors; \cite{wang2013duplicate} relies on the number of overlapping visual words; and \cite{kim2015near} uses the Jaccard similarity of the sets of visual words of two images.  Because of the limitations of the global and local descriptors, \cite{kim2015near} has encountered noisy clusters, which limits its application in recommendation systems.  Geometric verification, e.g. RANSAC \cite{fischler1981random}, greatly improves the precision of the classifier \cite{wang2013duplicate}, but is very computationally expensive on a web-scale dataset.  In this work, we collect human curated near-dupe criteria, and train an embedding-based deep near-dupe classifier to replace RANSAC in order to achieve high precision efficiently. Compared with existing web-scale efforts, we improve the design such that our system is capable of incremental updates.

% % % % % % % % % % % % %
%                            GOOG       MSFT2013    MSFT2015
% feature representation:   global       global      local
% candidate generation  : nn search       hash        hash
% neardup criteria      :   global        visual words overlap > ...  visual words overlap > ...

\section{System Overview and Evolution}
\label{sec:system-overview}

% % % % % % % % % % % % %
% Evolution of our system
\begin{table*}[!ht]
\centering
 \caption{Evolution of the Near-dupe system over time.}
  \vspace{-.3cm}
 \begin{tabular}{c c c  c  c} 
  \toprule
      & \multicolumn{1}{p{4cm}}{\centering First Generation (Mid 2013)} 
      & \multicolumn{1}{p{4cm}}{\centering Second Generation (Early 2014)} 
      & \multicolumn{1}{p{4cm}}{\centering Third Generation (Mid 2018)} \\
  \midrule
      Applications
      &  De-duplicate search feed
      & \multicolumn{1}{p{4cm}}{\centering Aggregate pin meta data \\ Remove copyrighted content } 
      & \multicolumn{1}{p{4cm}}{\centering Canonical pin selection \\ De-duplicate home feed}  \\ 
  \midrule
      Requirements 
      & \multicolumn{1}{p{4cm}}{\centering Medium precision \\ Medium recall} 
      & \multicolumn{1}{p{4cm}}{\centering High precision \\ Medium recall} 
      & \multicolumn{1}{p{4cm}}{\centering High precision \\ Medium-high recall \\ High efficiency} \\
 \midrule
      Image Representation & Global features & Bag-of-visual-words (BOW) & Visual embedding \\
      Candidate Generation & LSH & Inverted index & LSH on Spark \\
      Candidate Selection &  RANSAC & RANSAC + Decision tree  & Deep classifier \\
      Clustering &  Map-Reduce & Map-Reduce & Greedy K-cut on Spark \\
 \midrule
      \multicolumn{1}{p{4.3cm}}{\centering Detected Near-dupes in \\Image Corpus (Includes Spam)} & 23\% & 46\% & 64\% \\
    
 \bottomrule
 \end{tabular}
 \label{system-evolution}
 \vspace{-.3cm}

\end{table*}
% System overview
The near-dupe image detection system is composed of three main components illustrated in Figure \ref{fig:system-diagram}: \textit{candidate generation} largely solves scaling and drives the recall, \textit{candidate selection} prunes the candidates to drive the precision, and \textit{clustering} approximates equivalence partitions over the universe of images.

% Previous Generations
Each component has evolved drastically as application needs change and with the emergence of newer approaches and technology. Table \ref{system-evolution} illustrates how each component evolves over time.

The first near-dupe detection system, developed in 2013, aimed to remove near-dupe pin\footnote{On Pinterest, a pin is created when a user bookmarks an image, and it contains meta data such as user-generated text description, board title, destination url, etc.} images on the search page for better user experience, as shown in Figure \ref{fig:search-dedupe}.  
In this system, images were represented by global image descriptors, and we used LSH-based search to generate near-dupe candidates.  At the candidate selection stage, we extracted local image descriptors and applied geometric verification using RANSAC as the near-dupe classifier.  The clustering stage implemented Algorithm \ref{transitive-closure} on Map-Reduce, detailed in Section \ref{method-clustering}.  The first system was efficient, but the drawback of global descriptors limited the recall and precision.
Since false positives did not affect user experience in the search result, the system did not require high precision.

% and we can combine it with non-image-based methods to remove near-dupe images in search results (such as de-dupe using pin , 

% \begin{figure}[htb]
% \begin{minipage}[b]{1.0\linewidth}
%   \centering
%   \centerline{\includegraphics[width=6.5cm]{figures/neardup-exp-dedupe-search.png}}
% \end{minipage}
% \caption{In the early days of Pinterest, search results for some queries suffer from massive number of near-dupe images. (a) Search results without image de-dupe (b) Search results after applying near-dupe signals.}
% \label{fig:search-dedupe}
% \end{figure}

As the volume of data grew significantly on Pinterest, new applications emerged and required higher precision, such as removing copyrighted images and aggregating pin meta data.  In order to overcome the limitation of global image descriptors, the second generation system in 2014 adopted the popular BOW image representation, which is ideal for retrieval.  We obtained near-dupe candidates by retrieving similar images from an inverted index containing all images.  To improve the precision of the near-dupe classifier, we employed a two-stage near-dupe classifier containing (1) geometric verification using RANSAC to reject false matches and (2) a decision tree model to guard against near-dupe criteria.  The decision tree model was trained using human-labelled data and used features such as differences in global image descriptors, size and location of the overlapping area of the two images after being geometrically aligned, etc.

The second generation had much better recall and precision, but it was inefficient in space and had high computational cost.  Leveraging the emergence of deep learning and Spark, the third generation focused on bringing the recall higher while increasing the efficiency of daily processing to keep up with the growth of the content corpus. The system was deployed in mid-2018. Section \ref{sec:methods} describes the design in more detail.

% -----------------------------------------------------------
\section{The Latest Generation}
\label{sec:methods}
In this section, we present each component shown in Figure \ref{fig:system-diagram}.

\subsection{Candidate Generation}
\label{sec:methods-cg}
\subsubsection{Image Representation as LSH Terms}
% In order to understand the content of an image, we map an image into an embedded vector space. Visual embeddings are high-dimensional vector representations of images which capture visual and semantic similarity. They are typically produced via neural network architectures like VGG16 \cite{simonyan2014very} and Inception \cite{szegedy2016rethinking}. 
In this system, each image is represented by the visual embedding from \cite{jing2015visual}.  To cluster images via the near-dupe relation, each day we compare tens of millions of new images to billions of existing clusters. Approaching such a nearest neighbor search without an optimization would yield quadratic time complexity and a runtime proportional to more than 10 quadrillion image comparisons. Instead, we use a reduction of visual embeddings into LSH terms to drastically improve the tractability of such a problem.

Through a process based on bit selection of binarized embeddings, we first reduce the dimensionality of the original embedding space. Next, the derived bits are grouped into numeric LSH terms in a process that fundamentally trades off detection probability and runtime. The smaller the groupings, the more computationally expensive it is to run the nearest neighbor search, but this increases the probability of accurate detection. Nearest neighbor search uses LSH terms and their Jaccard overlap as an approximation of cosine similarity between vectors in the original embedded space.

\subsubsection{Batch LSH Search}
With every image being represented by a set of LSH terms, we build an inverted index and implement a batch-oriented search over images. It is important to note that we only search against images that represent the entire cluster (known as cluster heads). The selection of cluster heads is a combination of engagement- and content-based heuristics at the time of initialization of the cluster. This allows to reduce the search space with minimal recall loss due to candidate augmentation, described in Section \ref{candiate_augmentation}. To make batch-oriented search efficient at a high level, we use a combination of functional transformations, compressed inverted indexes and joins to calculate a result set for all the query images at once. The batch LSH search is implemented on Spark and requires a series of optimizations, such as dictionary encoding \cite{10.5555/561620} and variable byte encoding \cite{Cutting:1989:ODI:96749.98245}, to ensure we can handle the data volumes even in a much more computationally tractable LSH term space. A combination of dictionary encoding and variable encoding on image id can reduce inverted index storage space by 20\%. Algorithm \ref{candiate_generation} details the transformations.

% \IncMargin{1em}
\begin{algorithm}
\SetKwProg{Line}{}{}{}
\SetKwInOut{Input}{Input}
\SetKwInOut{Output}{Output}

\Input{
$RDD[(indexImgId, List[LSHTerm])]$ \tcp*{index} \\
\nonl$RDD[(queryImgId, List[LSHTerm])]$ \tcp*{query}
}{}
\Output {$RDD[(queryImgId,TopK[(indexImgId,Overlap)])]$}{}

\nonl   \Line{
\tcp*[l]{create inverted index for query \& index}
flatMapValues, groupBy query $\xrightarrow{} RDD[(LSHTerm, List[queryImgId])]$  \\
flatMapValues, groupBy index $\xrightarrow{} RDD[(LSHTerm, List[indexImgId])]$
}{}
\nonl   \Line{\tcp*[l]{search results by query}
Join (1) and (2), flatMap over queries posting list, groupBy query 
$\xrightarrow{} RDD[(queryImgId, List[List[indexImgId])]$ ;
}{}
\nonl   \Line{\tcp*[l]{count number of times each ImgId is seen}
Merge $List[List[indexImgId]]$ into $TopK(indexImgId, Overlap)$}{}
\caption{Candidate Generation on Spark}
\label{candiate_generation}
\end{algorithm}
% \DecMargin{1em}

\subsubsection{Candidate Augmentation}
\label{candiate_augmentation}
A key observation is that if the query image does not match the cluster head via a trained classifier, it may match some other images in the cluster - even those that are extremely similar to the cluster head. To illustrate this, assume a cluster $A \longleftarrow \{B:99,C:99,D:70\}$, where $A$ is the cluster head, the similarity scores with respect to $A$ are as listed, ranging from 0 to 100. A query image $Q$ may have scored relatively poorly with cluster head $A$, i.e. $sim(A,Q) = 10$, which is below a chosen $threshold$. We have observed that $sim(B,Q)$ or $sim(C,Q)$ may be above the $threshold$ even though $B$ and $C$ are extremely close to $A$ visually. The combination of $(A,Q)$ presents a difficult or adversarial example for the trained classifier in Section \ref{sec:methods-cs}, but other images in the cluster may not necessarily share that characteristic. Top augmentation images that are close to the cluster head (e.g. $B$, $C$) are added to the candidate list and increase end-to-end recall of the system by $13\%$.

\subsubsection{Bootstrapping Over High Precision}
\label{boostrapping_high_precision}

A number of iterations of the improvements in the near-dupe system have relied on data produced in a previous iteration and the property of near-exact precision. Candidate augmentation is also heavily reliant on near-exact precision of both the classifier and the consistency of the existing clusters. We are able to consider $Q$ as matching cluster $A$ when any of top $k$ near-dupes in $A$ match $Q$. This can only be possible when clusters are consistently precise. While the near-dupe relation is not transitive, in general by keeping the classifier at near-exact precision we are able to make that assumption for close neighbors.

\subsection{Candidate Selection}
\label{sec:methods-cs}

\subsubsection{Near-dupe Classifier}
Batch LSH is a powerful way to generate high recall while minimizing computational cost. However, it will generally not produce optimum precision and ranking of candidates. We use a pass over generated candidates via a supervised classifier to select the candidates that are similar enough to be considered near-dupes. 

\begin{figure}[htb]
\begin{minipage}[b]{1.0\linewidth}
  \centering
  \centerline{\includegraphics[width=8cm]{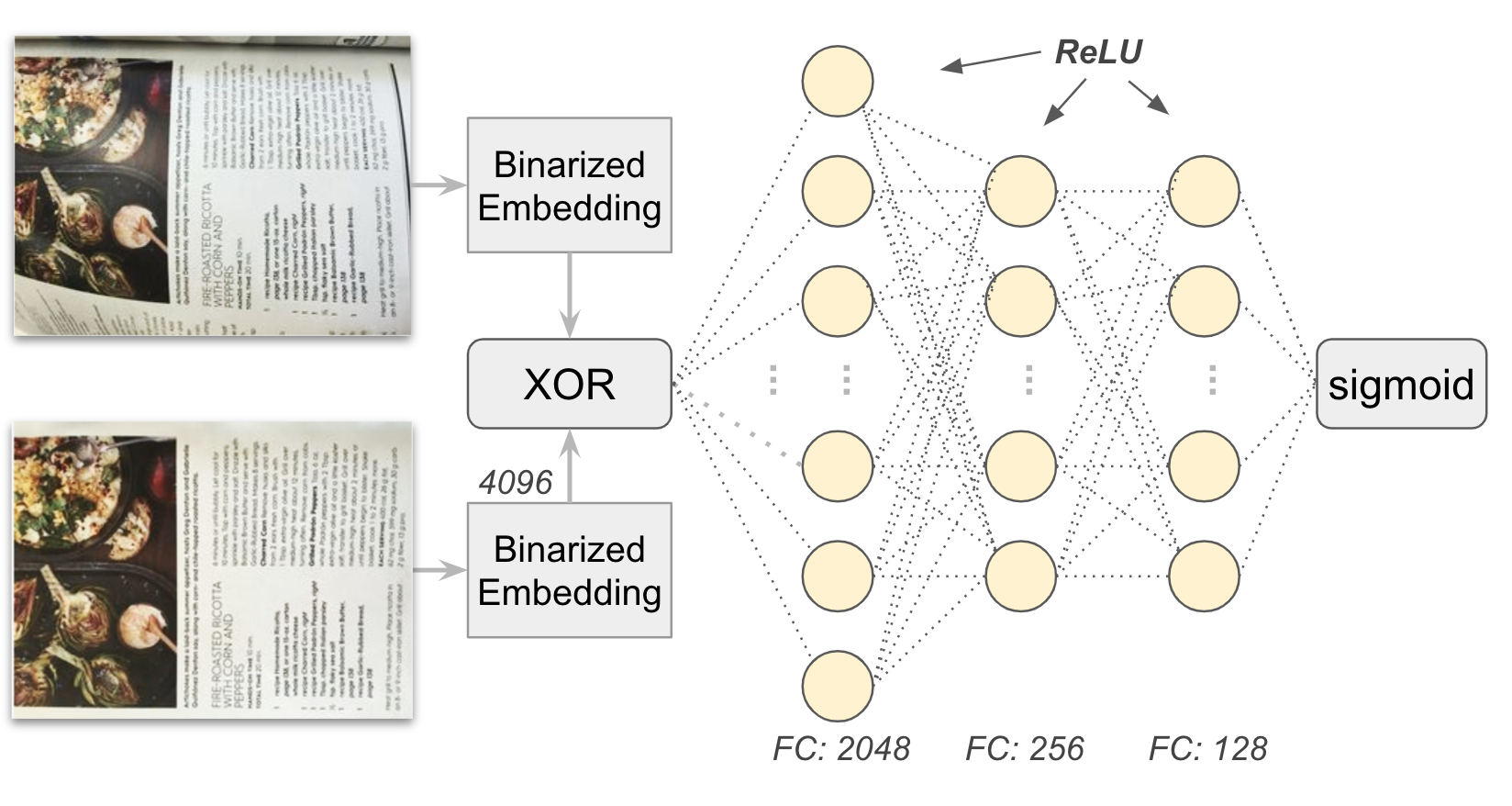}}
\end{minipage}
  \vspace{-.6cm}
\caption{Architecture of the near-dupe classifier.}
\label{fig:classifier-diagram}
\vspace{-.3cm}
\end{figure}

The task of the classifier is to select near-dupe images from the candidates produced by LSH and the augmentation.  Figure \ref{fig:classifier-diagram} illustrates how this classifier works.  To improve learning and convergence for each pair of images, hamming bits (binarized visual embeddings from \cite{jing2015visual}) are XORed and fed into the input layer, followed by 3 layers of fully connected layer and a sigmoid.

The classifier is implemented in TensorFlow with Adam optimizer \cite{kingma2014adam} over a set containing more than one billion distinct pairs described in Section \ref{classifier-training-data}. 
% The training set was derived from the output of a decision tree classifier over SURF visual features with geometric verification, which was used in previous generation of the near-dupe system. 
% The classifier is tuned for high precision and achieves over 99\% precision on human-labeled data. 
% To further boost the performance the classifier is fune-tuned on human labeled evaluation data.
The inference over the trained network happens in the Spark context to efficiently vectorize and reduce overhead. With a network that has almost 10 million parameters, we achieve an average of 1-2ms per prediction on a cluster of r5d machines on Amazon Web Service. 
\label{method-cs-xor}

\subsubsection{Training Labels for the Classifier}
\label{classifier-training-data}
In order to collect a large amount of training labels for the near-dupe classifier, we used detection results from the previous generation of the near-dupe detection system over LSH candidates as well as labels derived from augmentation as described in Section \ref{cs:augmentation_data}.  The previous generation classifier is detailed in Section \ref{sec:system-overview}.

% The previous system employs SURF features \cite{bay2006surf}, a fast local feature descriptor, to represent each image, and a two-stage near-dupe classifier containing (1) geometric verification using RANSAC \cite{fischler1981random} to reject false matches and (2) a decision tree model to guard against our definition of near-dupe, i.e. heavy-cropping does not meet our definition of near-dupe.  
% %Even when the pair passes geometric verification, they are not necessarily near-duplicate by our definition.  
% % We built a decision tree model with a depth of 8 using CART \cite{breiman2017classification}.  
% The decision tree model uses features such as the overlapping ratio of the two images after being geometrically aligned, mean differences in each color channel, etc.  %We collected ~1000 pairs (1:1 positive-to-negative example ratio) and% trained a decision tree model with a depth of 8 using CART \cite{breiman2017classification}.

\subsubsection{Augmentation Labels}
\label{cs:augmentation_data}
Following the Augmentation procedure described in Section \ref{candiate_augmentation}, we generate a new training set for the next generation classifier that specifically targets adversarial examples. In the provided example, note that even though $Q$ was (correctly) recorded into cluster $A$ due to the transitivity over $B$, the classifier was not able to handle $sim(Q,A)$ correctly. Learning over such pairs is an effective way to increase accuracy.
\label{method-cs-dt}

\subsection{Clustering}
\label{method-clustering}

\begin{figure}[b]
\begin{minipage}[b]{1.0\linewidth}
  \centering
  \centerline{\includegraphics[width=6cm]{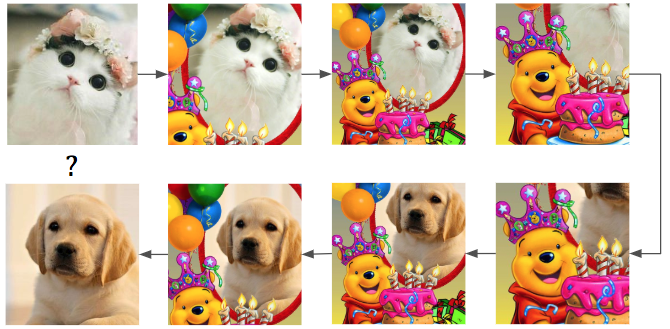}}
\end{minipage}
\vspace{-.6cm}
\caption{An image of a cat can morph into an image of a dog through near-dupe hops.}
\label{fig:neardup-cluster-wrong}
\vspace{-.3cm}
\end{figure}

Finding an ideal partition over the universe of images is mathematically not well-defined, because the near-dupe relation is not transitive and therefore not an equivalence relation.  For example, an image of a cat gradually morphs into an image of a dog as shown in Figure \ref{fig:neardup-cluster-wrong}. Each iteration would fall well within the near-dupe criteria, yet it is not clear how to partition the sequence. 

% \IncMargin{1em}
\begin{algorithm}
\SetAlgoLined

 \SetKwFunction{Concat}{Concat}
 \SetKwFunction{Min}{Min}
 \SetKwBlock{Initialization}{Initialization}{}
 \SetKwFunction{emit}{Emit}

 \SetKwProg{Map}{Map}{}{}
 \SetKwProg{Reduce}{Reduce}{}{}
 \SetKwInOut{Input}{Input}
 \SetKwInOut{Output}{Output}
 
 \Input{Edges $(img\_id1, img\_id2)$}
 \Output{Transitive closure clusters $C$}
 \Initialization{
     \Map{edge ($img\_id1, img\_id2$)} {
        $edge\_id \leftarrow \Concat(img\_id1, img\_id2)$\;
         \emit($img\_id1, img\_id2, edge\_id$)\; 
     }
 }
 \While{$num\_edge\_updated > 0$}{
  \Map{$(img\_id1, img\_id2, edge\_id)$}{
    \emit ($img\_id1, img\_id2, edge\_id$)\;
    \emit ($img\_id2, img\_id1, edge\_id$)\;
  }
  \Reduce{on the first element of the tuple}{
   $group\_id \leftarrow \Min(edge\_id)$ in the group\;
   $num\_edge\_updated \leftarrow 0$\;
     \ForEach{$(img\_id1, img\_id2, edge\_id)$}{
        \If{$edge\_id \ne group\_id$}{$num\_edge\_updated$ += 1}
        \emit ($img\_id1, img\_id2, group\_id$)\; 
     }
  }
 }
 
 \caption{Transitive Closure Algorithm}
 \label{transitive-closure}
\end{algorithm}
% \DecMargin{1em}

We represent images in a graph, where nodes are the images and edges represent image similarity.  We first apply the transitive closure algorithm (Algorithm \ref{transitive-closure}) over coherent sub-clusters. Each sub-cluster is a set of images that pass the near-dupe threshold for a given query image. 

% We use a combination of transitive closure over coherent sub-clusters (illustrated in Algorithm \ref{transitive-closure}) and a greedy k-cut to find an approximation to the partition that minimizes the k-cut over the graph. 

% \subsubsection{K-cut}
Following transitive closure, we greedily perform k-cut to correct for any errors that transitive closure may have introduced over several sub-clusters, illustrated in Algorithm \ref{algo-k-cut}. We rely on the fact that transitive closure was formed from coherent sub-clusters to develop a greedy approximation algorithm that is tractable.

\subsection{Incremental Update Algorithm}
We design our algorithm to handle a new batch of images on a daily basis.  The algorithm works in two stages.  In the first stage, we use the new images to query against (a) existing image clusters (NvO job) and (b) new images (NvN job), shown in Table \ref{NvONvN}.  We only run transitive closure for (b) since we want to recognize the new image cluster within the new images.  In the second stage, we merge the near-dupe image mappings from the NvO job and the NvN job, with the preference of NvO over NvN. Members of the NvN cluster will be mapped to the old cluster if such a match is found. Otherwise the NvN cluster will be added as a new cluster. This enables us to continue to grow existing image clusters.
% For example, if the result from (a) contains $A \xrightarrow{} B$, and results from (b) contains $A \xrightarrow{} C$, the finally mapping would be $A \xrightarrow{} B$.  

\begin{table}[!ht]
\centering
 \caption{Two-stage approach to process new images. (a) NvO job: use new images to match existing images (b) NvN job: use new images to match new images. TC: transitive closure.}
   \vspace{-.3cm}
\begin{tabular}{c | c  c  c} 
 \toprule
 Name & Query set & Index set & TC + K-cut \\
 \midrule
 (a) NvO job & New images & Existing images & No \\
 (b) NvN job & New images & New images & Yes  \\
 \bottomrule
 \end{tabular}
\label{NvONvN}
 \vspace{-.4cm}
\end{table}

% -----------------------------------------------------------
\section{Experiments}
\label{sec:experiments}
\subsection{Gold Standard Dataset}
\label{sec:gold-standard-dataset}

% \IncMargin{1em}
\begin{algorithm}
\SetAlgoLined

 \SetKwFunction{Concat}{Concat}
 \SetKwFunction{Min}{Min}
 \SetKwFunction{Classify}{Classify}
 \SetKwFunction{rand}{Rand}
 \SetKwInOut{Input}{Input}
 \SetKwInOut{Output}{Output}
 
 \Input{Transitive closure clusters $C^{(i)} \in C, i = 1..|C|$}{}
 \While{$C \ne \emptyset$}{
    \ForEach{cluster $C^{(i)} \in C$}{
        $h \leftarrow \rand(|C^{(i)}|), C^{'} \leftarrow \{\}$\;
        \ForEach{image $C^{(i)}_{k} \in C^{(i)}, k \ne h$}{
            \If{$\Classify(C^{(i)}_{h}, C^{(i)}_{k}) > t$}{
                $C^{'} \leftarrow C^{'} \cup {C^{(i)}_{k}}$\;
            }
        }
        \If{$C^{'} \ne \emptyset$}{
            $C^{(i)} \leftarrow C^{(i)} \setminus ({C^{(i)}_{h}} \cup C^{'})$\;
        }
        $C \leftarrow C \cup C^{'}$\;
    }
 }
 \caption{Greedy Approximate K-cut Algorithm}
 \label{algo-k-cut}
\end{algorithm}

We created a dataset\footnote{Available at \url{https://www.pinterest.com/pin/100275529191933620}.} of \textasciitilde53,000 pairs of images by using a crowdsourcing platform (86\% non-near-dupe and 14\% near-dupe image pairs).  The dataset was evaluated by human judges who have achieved high accuracy against a smaller golden set.  For each pair, 5 different human judges were asked whether the pair constitutes near-dupe images.  To the best of our knowledge, there is no prior work that has created a human-curated dataset of this scale.

\subsection{Classifier}
\label{sec:classifier-results}
We compared the performance of our near-dupe classifiers with the one in the most recent published web-scale system \cite{kim2015near}, by evaluating on the gold standard dataset.
We implemented the near-dupe classifier based on the Jaccard similarity between the visual words described in \cite{kim2015near}.

The \textit{Decision tree classifier} is described in Section \ref{sec:system-overview}. The \textit{Base deep classifier} described in Section \ref{method-cs-xor} is trained with a dataset containing 100 million pairs of images generated based on the approach detailed in Section \ref{classifier-training-data}.

We further improved the classifier performance by refining the training set. The \textit{Production deep classifier} differed from the base deep classifier in increasing the size, diversity, and difficulty of the training examples to over 1 billion pairs of images. Table \ref{classifier-results} lists the precision-recall (PR) and ROC AUC of the above classifiers.  By comparing the performance of the classifiers using various image features, the visual embedding and the amount of difficult training data play a big role in improving the performance significantly.

In the latest generation system, we chose an operating point that maximizes precision while having medium-to-high recall\footnote{One application of the system is to remove copyrighted content by finding near-dupe copies of the content.  During one operation, human judges reviewed 6,700 detected near-dupe copies of one copyrighted image, and found 0 false positive.}.

\begin{table}[!ht]
\centering
 \caption{Classifier performance on the gold standard dataset.}
  \vspace{-.3cm}
 \begin{tabular}{c  c  c} 
 \toprule
  &  PR AUC & ROC AUC \\
 \midrule
  Kim et al. \cite{kim2015near}    & 0.58 & 0.83 \\
  RANSAC + Decision Tree (Gen. 2)    & 0.64 & 0.87 \\
  Base Deep Classifier         & 0.80 & 0.96 \\
  Production Deep Classifier (Gen. 3)   & \textbf{0.85} & \textbf{0.97} \\
 \bottomrule
 \end{tabular}
 \label{classifier-results}
 \vspace{-.5cm}

\end{table}

\subsection{Cluster Pinterest Images}
\begin{figure}
  \begin{subfigure}[b]{0.6\columnwidth}
    \includegraphics[width=6cm]{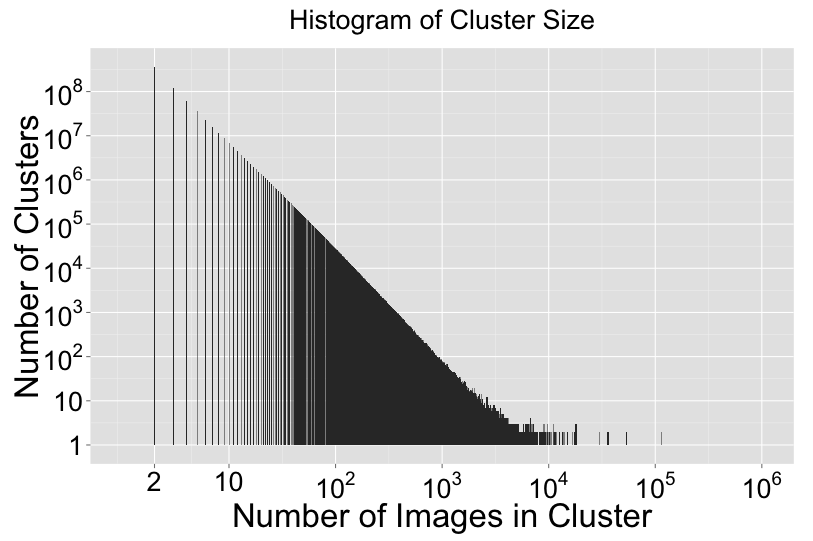}
    \centering
  \vspace{-.6cm}
    \caption{}
    \label{fig:cluster-histogram}
  \end{subfigure}
  \hfill %%
  \begin{subfigure}[b]{0.23\columnwidth}
    \centering
    \includegraphics[width=\linewidth]{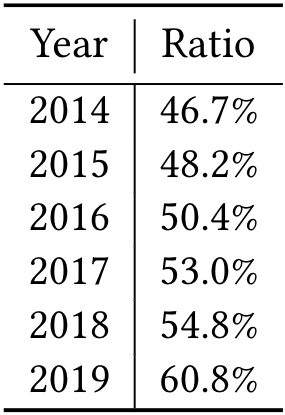}
    \textcolor{white}{\rule{1cm}{0.4cm}}
%   \vspace{-.6cm}
    \caption{}
    \label{fig:neardupe-ratio}
  \end{subfigure}
  \vspace{-.45cm}
  \caption{(a) Distribution of near-dupe cluster sizes on Pinterest dataset. (b) Increasing trend of the percentage of non-spam images created on Pinterest that are near-dupe.}
  \label{fig:exp-pinterest-results}
% \vspace{-.5cm}
\end{figure}

We applied the proposed method to a dataset of over 8 billion images crawled from Pinterest, that comes with the visual embedding from \cite{jing2015visual}.  We found \textasciitilde900 million clusters, containing \textasciitilde5.4 billion (\textasciitilde64\%) near-dupe images\footnote{Prior web-scale works detected \textasciitilde13\%, 27\%, 34\% in their image corpus \cite{liu2007clustering,wang2013duplicate,kim2015near}.}.  The most common cluster size is 2, and average cluster size is 6.  The resulting distribution of near-dupe cluster sizes meets the power law distribution, as shown in Figure \ref{fig:cluster-histogram}.   

% Google, 200 million out of 1.5 B
% MSFT, 550 out 2B
% From the 1B images, we discovered about 82.2M clusters, which contain about 344.7M near-duplicate images

% Figure \ref{fig:neardup-top-images} shows some near-dupe clusters with large number of images.

We also conducted a longitudinal study on the percentage of near-dupe images on Pinterest by image creation time, and observed an increasing near-dupe ratio over time, in Table \ref{fig:neardupe-ratio}.  

% https://datahub.pinadmin.com/datahub/data_doc/47979/?cellId=1155000&executionId=1018671
%   692,305,266 - distinct canonical_sig
% 2,856,895,615 - distinct sig

\section{Applications at Pinterest}
\label{sec:applications}
The proposed near-dupe image detection system processes all Pinterest images and handles new images on a daily basis.  We show that this system is helpful in many applications on Pinterest \cite{jing2015visual,eksombatchai2018pixie, hamilton2017inductive,mao2016training}, where over 10 billion recommendations are made a day.  
% The lift in user engagement is summarized in Table \ref{tab:pinterest-lift}.
Table \ref{tab:pinterest-lift} summarizes the lifts in user engagement (e.g. pin click/save) in A/B experiments, after using the near-dupe detection results.

\textbf{De-duplicate Pinterest Feeds.}
The near-dupe mapping can be used to filter out duplicated content in search result pages and the user home feed, illustrated in Figure \ref{fig:search-dedupe}.  

% moved to experiment - pinterest
% \begin{table}[!ht]
% \centering
%  \caption{\% of images created on Pinterest platform that are duplicate by year}
%  \begin{tabular}{c | c} 
%  \toprule
%  Year & Ratio \\
%  \midrule
%  2014 & 46.7\% \\
%  2015 & 48.2\% \\
%  2016 & 50.4\% \\
%  2017 & 53.0\% \\
%  2018 & 54.8\% \\
%  2019 & 60.8\% \\
%  \bottomrule
%  \end{tabular}
%  \label{neardup-trend-by-year}
% \end{table}

\textbf{Aggregate Pin Meta Data.}
Pin meta data (e.g. descriptions, board titles) are aggregated among pin images from the same near-dupe cluster, and these meta data are used in generating keywords for search engine optimization (SEO) and serving pin recommendations on Pinterest.  For graph-based recommendation engines \cite{eksombatchai2018pixie, hamilton2017inductive}, images within the same near-dupe cluster are condensed into one node, and this reduces the size of the graph by $2.5x$ and greatly increases the connectivity of the graph.

\textbf{Canonical Pin Selection.}
The near-dupe system can be leveraged to select the most appropriate pin based on the context of users. Given a cluster of near-dupe images, each image may link to a different pin.  For example, recommendation systems can select the most localized content in the selected near-dupe cluster.

\begin{table}[!ht]
\centering
 \caption{Summary of engagement lift by using near-dupe mapping, measured through A/B experiments on Pinterest.}
  \vspace{-.4cm}
  \begin{tabular}{c  c} 
 \toprule
 Applications & Engagement Lift \\
  \midrule
 Volume in SEO Traffic & +33\% \\
 De-duplicate Search feed  & +11\% \\
 De-duplicate Home feed  & +7\% \\
 Canonical Image Selection & +9\% \\
 \bottomrule
  \vspace{-.5cm}
 \end{tabular}
\label{tab:pinterest-lift}
  \vspace{-.5cm}
\end{table}

% -----------------------------------------------------------
\section{Lessons}
\label{sec:discussions}

\begin{figure}
  \begin{subfigure}[b]{0.48\columnwidth}
    \includegraphics[width=\linewidth]{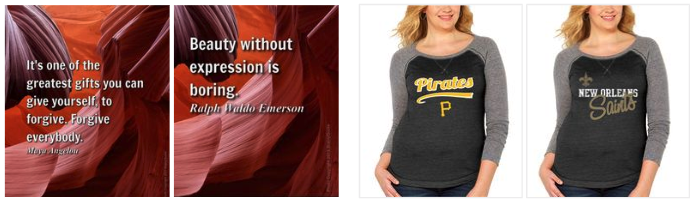}
  \vspace{-.5cm}
    \caption{}
    \label{fig:example_false_positive}
  \end{subfigure}
  \hspace{.05cm} %%
  \begin{subfigure}[b]{0.48\columnwidth}
    \includegraphics[width=\linewidth]{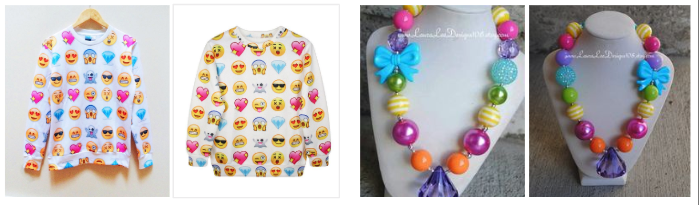}
  \vspace{-.5cm}
    \caption{}
    \label{fig:example_false_negative}
  \end{subfigure}
  \vspace{-.5cm}
  \caption{(a) Ambiguous image pairs. (b) Near-dupe pairs that were incorrectly classified by earlier classifier.}
  \label{fig:hard_examples}
  \vspace{-.5cm}

\end{figure}

\textbf{Near-dupe Criteria.}
It is hard to define the exact criteria of near-dupe, and we rely on a human-curated gold standard dataset to capture this, since our system aims to be used in recommendation systems.  However, human judges have noisy votes - for example, image pairs in Figure \ref{fig:example_false_positive} are classified as near-dupe by our system but are labeled as non-near-dupe by human judges.  Since images with text overlay often provide  little value in the recommendation system, such false positives are minor.

\textbf{Image Representation.}
We find that the semantics of the image is crucial to near-dupe classification.  Global and local descriptors from previous systems are pixel-based, while the more powerful visual embedding contains both pixel- and semantic-level signals and best mimics how humans visualize and identify near-dupe pairs.   Image pairs shown in Figure \ref{fig:example_false_negative} are mis-classified by the previous classifier due to the limitation of the pixel-based descriptors, and are correctly classified by the latest system.   The high precision of the system is key to enable the real-world applications in Section \ref{sec:applications}.

\textbf{Design Choice: LSH.}
For near-dupe detection task, a key metric for measuring the efficacy of LSH candidate generation is not $recall@k$ but rather $recall@distance$. Only neighbors that are visually close enough can be considered near-dupe; therefore, the aim is to efficiently find neighbors that fall within a given distance rather than to find all $k$ closest neighbors regardless of distance. We evaluated LSH recall by classifier selection on 5 billion LSH candidates produced by the near-dupe system; recall of candidate generation is over 99.99\% - less than 0.01\% of candidates outside of what LSH search generates would be detected as near-dupes by the classifier.  Thus, LSH search works well in the near-dupe system.

\textbf{Synthetic Labels.}
By optimizing for precision in iterations of this system, we were able to build a new generation of the system by bootstrapping over synthetic training labels. The third generation is trained on a large number of examples labeled by the previous classifier. Section \ref{sec:classifier-results} shows that a combination of synthetic training labels and powerful embedded representation leads to improvements in accuracy characteristics and ability to generalize.

% -----------------------------------------------------------
\section{Conclusion}
\label{sec:conclusion}
% The ability to partition an image space into near-dupe clusters allows downstream systems to greatly improve relevance and density of content. In this paper, we have described a multi-stage system that maintains high accuracy as well as scales to over 7 billion images. 
% In future iterations of the system, we would like to apply this methodology to video content. An important limitation of our approach is the dependency on the embedding, which may not preserve all the details needed for near-dupe detection. Therefore, we would also like to explore augmenting embedded representations to be aware of the near-dupe task. 

This paper presents the evolution of a web-scale near-dupe image detection system, and shares how we scale it to over 8 billion images with high precision, e.g. the use of visual embedding to decrease computational load for candidate generation and selection, improvement of the near-dupe classifier using synthetic labels, efficient implementation on Spark, and design for incremental update.  We also showcase how this system allows recommendation systems to greatly improve relevance and density of content.  
% In future iterations of the system, we would like to apply this methodology to video content.  
By sharing our experiences, we hope real-world applications can benefit from de-duplicating and condensing image content on the Web.

% An important limitation of our approach is the dependency on the embedding, which may not preserve all the details needed for near-dupe detection. Therefore, we would also like to explore augmenting embedded representations to be aware of the near-dupe task. 

% 1. Use of previous generation of system as synthetics labels because of optimizing for precision
% 2. Use of embeddings decreases computational load for both candidate generation as well as candidate selection
% 3. Structuring the system as incremental allows for tackling the scale

% -----------------------------------------------------------
\begin{acks}
We would like to thank Vitaliy Kulikov, Jooseong Kim, Peter John Daoud, Qinglong Zeng, Andrew Zhai, Matthew Fang, Raymond Shiau, Charles Gordon, Siyang Xie, Chao Wang, Zhuoyuan Li, Yushi Jing, and Jacob Hanger for their help throughout the work.
\end{acks}

%\vfill
%\pagebreak

\bibliographystyle{ACM-Reference-Format}
\bibliography{citation}

\end{document}